\documentclass{article}
\usepackage[utf8]{inputenc}
\usepackage{amsmath, amssymb}
\usepackage{graphicx}
\usepackage{authblk}
\usepackage{hyperref}
\usepackage{caption}
\usepackage{geometry}
\title{C2G-KD: PCA-Constrained Generator for Data-Free Knowledge Distillation}
\author[1]{Magnus Bengtsson}
\affil[1]{Department of Engineering, University of Borås}
\author[2]{Kenneth Östberg}
\affil[2]{Senior Developer at XXXYYY, Computer Engineering and Data Science}
\date{}

\begin{document}

\maketitle

\begin{abstract}
We introduce \textbf{C2G-KD}, a data-free knowledge distillation framework where a class-conditional generator is trained to produce synthetic samples guided by a frozen teacher model and geometric constraints derived from PCA. The generator never observes real training data but instead learns to activate the teacher's output through a combination of semantic and structural losses. By constraining generated samples to lie within class-specific PCA subspaces estimated from as few as two real examples per class, we preserve topological consistency and diversity. Experiments on MNIST show that even minimal class structure is sufficient to bootstrap useful synthetic training pipelines.
\end{abstract}

\section{Introduction}
Training deep neural networks typically requires large datasets, which may not be available in privacy-sensitive or resource-constrained domains. Data-free knowledge distillation (DFKD) \cite{chen2019data} has emerged as a promising approach where a student model learns from synthetic data generated via a pretrained teacher network. However, existing DFKD methods often fail to ensure structural alignment between synthetic and real data.

We propose \textbf{C2G-KD}, a method leveraging PCA-derived constraints to guide a conditional generator in producing class-specific synthetic samples without direct access to real data. Central to this approach is the use of PCA \cite{pearson1901pca} to impose topological constraints from minimal real samples, ensuring that generated images structurally align with class manifolds.

\paragraph{On the Semantics of Topology \cite{tenenbaum2000global,carlsson2009topology,moor2020topological,peirce1931collected,merleauponty1945phenomenology} and the Syntax of Typology \cite{wiener1948cybernetics,shannon1948mathematical,croft2002typology,dreyfus1972what}}

In most machine learning literature, data is treated as a syntactic object: vectors, tensors, labels. The model’s typology  is typically defined by a fixed set of class labels that it is trained to predict, and learning is formalized as mapping from structureless inputs to one of these predefined categories.

Yet this framing obscures a fundamental inversion. The physical world—be it handwritten digits, human gestures, or visual scenes—manifests first as \textbf{topological structures}: spatial forms that arise directly from physical observation. These forms are \textit{semantic in themselves}—meaning-bearing prior to any classification. They exist as observable configurations that require no linguistic label to be recognized or understood.

Typology, in contrast, emerges when we assign discrete labels to these continuous forms. It is a \textbf{syntactic construction}, created when we overlay an algorithmic naming system—such as a softmax classifier—on top of the underlying structure. In this light, typology does not reveal meaning; it imposes a rigid symbolic vocabulary onto a rich and continuous semantic field.

This distinction becomes critical when training models without access to real data. A classifier trained to predict a fixed set of labels can only recognize what it has been explicitly told exists. It cannot adapt to novel forms or reinterpret topology unless its typology is retrained. Conversely, a generator constrained by structural priors—such as PCA components derived from class-averaged shapes—can explore semantically meaningful variations without relying on labels at all.

In this work, we adopt a \textbf{topology-first} perspective. The generator is guided by structural constraints derived from PCA decomposition of polar-transformed data, not by labels. It explores a latent space shaped by the morphological essence of a class, learned from minimal real examples.

However, manipulating PCA values—while syntactically valid—does not in itself yield recognizable semantic identity. A projected sample may lie perfectly within the geometric structure of a class manifold, but it does not declare “I am a five” until it is interpreted through the lens of a typological system. In our framework, this role is played by a frozen teacher model, which serves as a \textbf{semantic discriminator}: it confirms whether the generated sample activates the correct class identity under an existing typology.

This separation between form and label—between semantic shape and syntactic assignment—allows us to treat generation and classification as distinct, yet coupled, processes. Only when topological plausibility is validated by typological recognition does a synthetic image attain both structural and semantic legitimacy.

\section{Class-Conditioned Generator for Knowledge Distillation (C2G-KD)}

This work introduces a data-free generation process where PCA analysis of polar-transformed images serves as a structural constraint guiding synthetic data generation. A class-conditional generator, trained alongside a frozen teacher model, learns to synthesize images whose essential morphology can be reconstructed using only a subset of principal components. This implies that the dominant structural variations of each class are captured by the leading PCA vectors.

By reducing data complexity—specifically, by projecting generated samples into low-dimensional PCA subspaces—the overall training process is simplified. The generator produces samples that are structurally aligned with real class manifolds, while the teacher ensures semantic correctness.

This framework supports a bottom-up view of model design, where the intrinsic structure of the data defines the necessary representational complexity. Rather than fitting arbitrary input data, the model adapts to a constrained latent structure induced by the morphology of the class itself.

The \textbf{teacher model} employed in this work is the classic \textbf{LeNet-5} architecture~\cite{lecun1998gradient}, a foundational convolutional neural network originally developed for handwritten digit recognition. Its proven effectiveness and compact design make it well-suited as a reference model in controlled experiments.

The \textbf{generator model} is a custom architecture, inspired by the principles of \textbf{Generative Adversarial Networks (GANs)} introduced by~\cite{goodfellow2014generative}. While not adversarial in training, our generator adopts the latent-to-image synthesis strategy pioneered in that work.

Figure~\ref{fig:enter-label} presents a schematic overview of the generator and the pretrained LeNet-5 teacher model.

\begin{figure}[ht]
    \centering
    \includegraphics[width=1\linewidth]{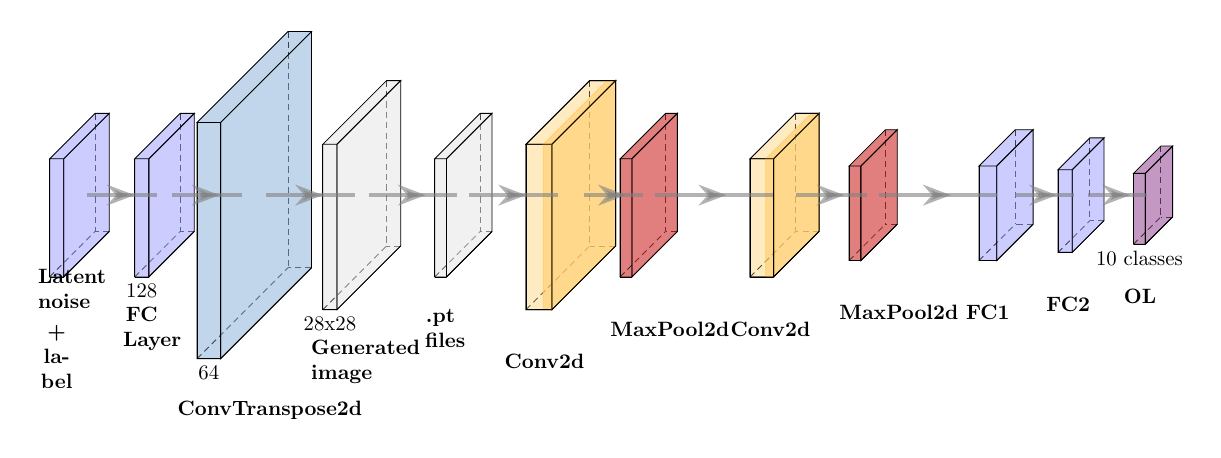}
    \caption{The generator model and the teacher model (pretrained LeNet-5).}
    \label{fig:enter-label}
\end{figure}
\subsection{Loss Functions}
The generator minimizes:
\begin{equation}
\mathcal{L}_{G} = \mathcal{L}_{\text{distill}} + \alpha \cdot \mathcal{L}_{\text{PCA}} + \beta \cdot \mathcal{L}_{\text{div}}
\end{equation}
Where:
\begin{itemize}
    \item $\mathcal{L}_{\text{distill}}$: Classification loss using teacher predictions.
    \item $\mathcal{L}_{\text{PCA}}$: Projection error to class PCA subspace.
    \item $\mathcal{L}_{\text{div}}$: Diversity term encouraging intra-class variation.
\end{itemize}

\subsection{Training Procedure}
Each class-specific generator is trained individually using PCA-guided loss and teacher supervision. Synthetic datasets are collected, optionally augmented, and used to train a student model.

\section{Theoretical Background PCA}

\subsection{Polar Sampling and Information Density}
Transforming images from Cartesian to polar coordinates introduces radius-dependent sampling density. Near the origin, sampling density tends toward infinity while spatial coverage collapses. At larger radii, sampling density per unit arc decreases. Mathematically, sampling density $D(r)$ satisfies:
\begin{equation}
D(r) = \frac{1}{r}
\end{equation}
This behavior parallels Shannon's self-information \cite{shannon1948mathematical}, where infinite sampling potential near the origin carries negligible information content.

This geometrical understanding informs our use of polar transformations in preprocessing and motivates localized dimensionality reduction strategies.

\subsection{Principal Component Analysis (PCA)}

\subsubsection{Worked Example: PCA on a Simple Matrix}
To illustrate the fundamental mechanics of PCA, consider a small dataset represented by a $3 \times 4$ matrix $X$:
\begin{equation}
X = \begin{bmatrix}
2 & 4 & 1 & 3 \\
3 & 5 & 2 & 4 \\
4 & 6 & 3 & 5
\end{bmatrix}
\end{equation}
Each column represents a variable (analogous to angular segments in polar images), and each row an observation.

\paragraph{Step 1: Centering the Data}
Subtract the mean of each column to center the data:
\begin{equation}
\bar{X} = X - \text{mean}(X) = \begin{bmatrix}
-1 & -1 & -1 & -1 \\
0 & 0 & 0 & 0 \\
1 & 1 & 1 & 1
\end{bmatrix}
\end{equation}

\paragraph{Step 2: Covariance Matrix}
Compute the covariance matrix $C$:
\begin{equation}
C = \frac{1}{n-1} \bar{X}^T \bar{X} = \begin{bmatrix}
1 & 1 & 1 & 1 \\
1 & 1 & 1 & 1 \\
1 & 1 & 1 & 1 \\
1 & 1 & 1 & 1
\end{bmatrix}
\end{equation}

\paragraph{Step 3: Eigen Decomposition}
Since the covariance matrix is uniform, the first principal component captures the entire variance. Solving $C v = \lambda v$, we find one dominant eigenvalue and its corresponding eigenvector $v_1$:
\begin{equation}
v_1 = \frac{1}{2} [1, 1, 1, 1]^T
\end{equation}

\paragraph{Step 4: Projection}
Project centered data onto $v_1$ to obtain principal component scores:
\begin{equation}
Z = \bar{X} \cdot v_1
\end{equation}
These scores represent the compressed representation of the dataset along its main direction of variance.

\paragraph{Step 5: Reconstruction}
To approximately reconstruct the data using the first component:
\begin{equation}
\hat{X} = Z \cdot v_1^T
\end{equation}
Then, adding back the column means yields an approximation of the original data.

This simple example demonstrates how PCA extracts dominant patterns (mean variation across columns) and allows reconstruction using a small number of components. In polar image analysis, columns correspond to angular segments, and this process reveals how segment relationships can be compactly described.

\subsubsection{Eigen-Digits PCA}
PCA applied across multiple images of the same class reveals dominant modes of variation, forming \textit{eigen-digits}. These capture class-specific variation patterns and allow visualization of global image structure. Projections into this subspace describe similarity between images.

\subsubsection{Radial-Segment PCA (Polar PCA)}
In polar-transformed images \cite{vonesch2015steerable} \cite{bengtsson2025pca}, PCA can be applied within a single image, analyzing angular variations across radial segments. This form of PCA captures local dependencies between angle segments within a digit. Unlike eigen-digits PCA, this method describes \textit{internal structure} rather than inter-image similarity.

Both PCA approaches underpin our generator’s constraints: eigen-digits PCA informs global class topology, while polar PCA reflects intra-image structural coherence.

\subsubsection{Theory}

Let the radial coordinate be denoted as $r$, and angular coordinate as $\theta$. Sampling along the angular dimension at a fixed radial distance $r$ divides the circle into equal angular steps $\Delta \theta$. The physical arc length corresponding to an angular step is given by:
\[
\Delta s = r \cdot \Delta \theta
\]

Therefore, the spatial resolution per angular sample is proportional to $r$:
\[
\text{Resolution per sample} \propto r
\]

The \textbf{sampling density}, defined as samples per unit arc length, becomes:
\[
D(r) = \frac{1}{r}
\]

As $r \to 0$, the sampling density approaches infinity:
\[
\lim_{r \to 0} D(r) = \infty
\]

This implies an infinitely high sampling potential at the center, though the spatial domain collapses into a single point.

\subsubsection{Information-Theoretic Analogy}

This behavior mirrors Shannon's concept of self-information:
\[
I(x) = -\log_2 p(x)
\]
where improbable events ($p(x) \to 0$) carry infinite information content.

Similarly, in polar sampling:
\begin{itemize}
    \item As $r \to 0$, sampling density $D(r) \to \infty$.
    \item Yet, this represents sampling over a single point, carrying trivial spatial information.
\end{itemize}

Thus, radial sampling near the origin exhibits an information paradox: infinite sampling density, yet negligible usable information.

In polar-transformed data, sampling density inherently depends on radius, following:
\[
D(r) = \frac{1}{r}
\]
This geometric property must be accounted for in image processing tasks to avoid over-sampling near the center and under-sampling at the periphery.

\subsubsection{PCA-Based Reconstruction and Halo Phenomenon}

In polar-transformed images, Principal Component Analysis (PCA) can be applied separately along radial segments. Each radial segment captures angular variations at a fixed radius, allowing PCA to extract the dominant patterns efficiently.

Because the angular resolution inherently depends on radius (as described previously), PCA can represent most of the essential image information using a relatively small number of principal components.

However, an important observation arises during image reconstruction using a limited number of PCA components: a visual \textbf{halo effect} often appears around the reconstructed object.

This halo phenomenon is directly linked to Shannon's information theory principles:
\begin{itemize}
    \item At larger radii, the available spatial sampling density decreases.
    \item This effectively reduces the \emph{local bandwidth} for information encoding.
    \item As a result, less information can be accurately captured or reconstructed at higher radii.
    \item The halo effect represents this \emph{loss of information} and manifests as noise or blur in outer regions.
\end{itemize}

In formal terms, the reduced local information bandwidth at large $r$ results in increased reconstruction error:
\[
\text{Reconstruction Error}(r) \propto r
\]

Thus, the halo is not merely an artifact, but a geometrically and information-theoretically expected consequence of radial sampling properties and bandwidth limitations.
\begin{figure}[h]
    \centering
    \includegraphics[width=1.0\linewidth]{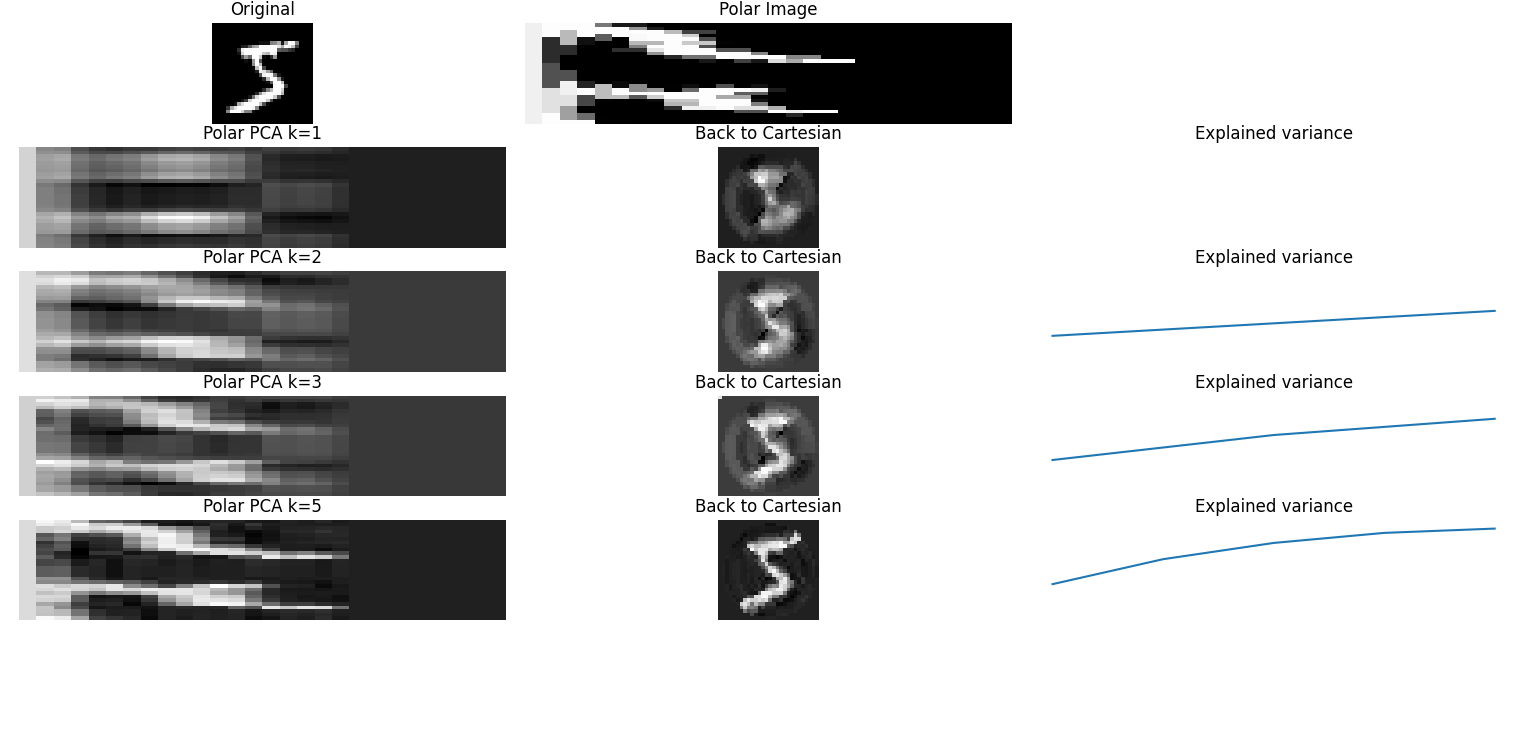}
    \caption{Reconstruction and the Halo Phenomenon}
    \label{fig:enter-label}
\end{figure}
\bigskip

In conclusion, PCA-based reconstruction of polar images inherently exhibits spatially varying fidelity, tied to the radial dependence of sampling density and information capacity.

\subsubsection{Methodological Approach: Bottom-Up Model Design}

This work adopts a \textbf{bottom-up} model design philosophy, where the structural properties of the data guide the formulation of both representation and learning. Rather than imposing a fixed model structure regardless of data characteristics, we emphasize a dynamic alignment between model complexity and data complexity.

A central insight is that model capacity should reflect—not exceed—the representational needs of the task. When the intrinsic geometry of the data is known or can be inferred, this knowledge can be used to design more efficient pipelines without sacrificing performance. This is not a critique of high-capacity models, but rather an affirmation that model complexity becomes most powerful when it is meaningfully matched to the structure of the input space.

In this study, we assume that the data possesses latent geometric regularities, which can be explicitly extracted and utilized. Our approach includes:

\begin{itemize}
    \item Transforming data from Cartesian to polar coordinates to exploit known symmetries.
    \item Analyzing sampling density and information content as functions of radius to guide local processing strategies.
    \item Applying dimensionality reduction techniques (e.g., PCA) in localized, structurally meaningful segments.
    \item Using classification strategies tailored to the resulting low-dimensional, structured representations.
\end{itemize}

This approach follows a principled path:

\begin{enumerate}
    \item Analyze and expose the structure of the data.
    \item Design representations that honor and preserve this structure.
    \item Choose learning mechanisms that complement the representational space.
\end{enumerate}

The result is a pipeline in which each component—preprocessing, representation, and classification—is informed by the geometry of the data itself. This enables improved performance, interpretability, and resource efficiency without compromising expressiveness.

\section{PCA-Constrained Training}
Using 2 real images per class, we compute a PCA subspace with 2 components. Generated samples are projected into this subspace to compute reconstruction error, forming the \textit{PCA Projection Loss}. This loss encourages structural similarity to real data without revealing exact image details.

\subsection{Synthesis of Class Variations via PCA Randomization and Teacher Validation}

By treating the class-average image as a structural prototype, a PCA decomposition of its polar-transformed representation can serve as a geometric prior. The principal components capture the dominant segment relationships characteristic of the digit class.

By deliberately randomizing the component weights within controlled limits, it becomes possible to generate new angular segment configurations that still adhere to the general class topology. Each randomization effectively produces a new variation—a synthetic "handwriting style"—that remains structurally valid but visually distinct.

However, geometric plausibility does not guarantee semantic correctness. Therefore, an external \textbf{teacher model} is used as a semantic validator: only those synthetic images that activate the target class in the frozen classifier are retained as valid samples.

This process conceptually separates \textbf{structure synthesis} (via PCA-based geometric variation) from \textbf{semantic validation} (via teacher feedback), enabling the generation of novel, class-consistent samples without direct data access. In effect, the PCA acts as a structural generator, while the teacher acts as a semantic discriminator.

\subsection{Class-Mean PCA Projection and Reconstruction}

Let $\{I_1, I_2, I_3, I_4\}$ be four MNIST images of the same digit class, each transformed into polar coordinates to obtain $\{P_1, P_2, P_3, P_4\}$. A class-mean polar image is computed as:
\begin{equation}
P_{\text{mean}} = \frac{1}{4} \sum_{i=1}^{4} P_i
\end{equation}

Principal Component Analysis (PCA) is then applied to $P_{\text{mean}}$, yielding a low-rank class-specific component base:
\begin{equation}
P_{\text{mean}} \approx Z \cdot W + \mu
\end{equation}
where $Z$ denotes the component scores, $W$ the PCA component matrix, and $\mu$ the mean vector.

For each individual image $P_i$, projection onto the class PCA base is computed as:
\begin{equation}
Z_i = (P_i - \mu) \cdot W^T
\end{equation}

Reconstruction of $P_i$ using the class-specific PCA base is then given by:
\begin{equation}
\hat{P}_i = Z_i \cdot W + \mu
\end{equation}

This reconstruction $\hat{P}_i$ represents the component of $P_i$ that aligns with the geometric structure of the class average. The reconstruction error thus measures the deviation of $P_i$ from the class-typical structure.

In practice, this process acts as a \textit{structural similarity projection}: only the class-consistent aspects of an input image are retained in its reconstruction, while non-conforming details are suppressed.

\subsubsection{Inverse Polar Transform and Reconstruction}

To visualize reconstructed polar images in a human-interpretable form, they are mapped back to Cartesian coordinates using the inverse polar transform. This operation remaps each pixel from polar coordinates $(r, \theta)$ to Cartesian coordinates $(x, y)$, centered at the image midpoint:
\begin{equation}
(x, y) = \text{PolarInverse}(r, \theta)
\end{equation}

In practice, the inverse transformation from polar to Cartesian coordinates requires careful numerical handling to preserve geometric fidelity. While libraries such as OpenCV offer built-in functions like \texttt{warpPolar} with the \texttt{WARP\_INVERSE\_MAP} flag, we found that this approach occasionally introduced visual artifacts and distortions in the reconstructed images—particularly in regions near the origin or along the angular wrap boundary.

To address these issues, we implemented a custom inverse transformation function that explicitly maps each polar coordinate \((r, \theta)\) back to its corresponding Cartesian coordinate \((x, y)\) using subpixel interpolation. This allowed us to control sampling behavior, avoid wrap-around artifacts, and preserve the structural integrity of the reconstructed images more reliably.

The resulting Cartesian images offer a clearer and more faithful approximation of the original digit geometry, particularly when evaluating reconstructions with a limited number of principal components.

This step is critical in our evaluation pipeline. Reconstructed polar images convey structural information across radial and angular segments, but are not visually meaningful until mapped back to Cartesian space. By inverse-transforming PCA-reconstructed polar images, we can directly compare them to real MNIST digits, both qualitatively and quantitatively.

\begin{figure}[h]
    \centering
    \includegraphics[width=1\linewidth]{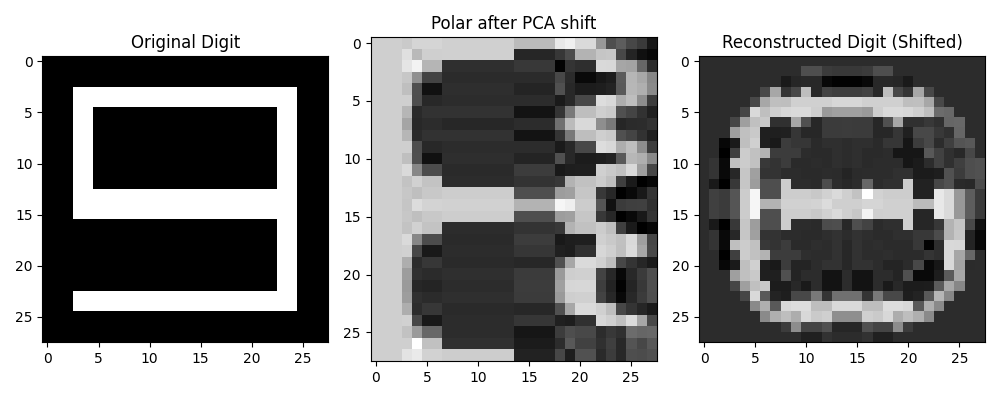}
    \caption{Enter Caption}
    \label{fig:enter-label}
\end{figure}
\subsection{Latent Morphological Representation via PCA in Polar Coordinate Space}

We propose that the morphological structure of digit classes can be effectively modeled as a latent shape space, where each instance is represented as a probabilistic linear combination of structural base components. Using the digit ``5'' as a representative example, we construct this latent space by applying Principal Component Analysis (PCA) to polar-transformed training samples.

Unlike Cartesian representations, the polar coordinate transform emphasizes radial and angular variations, thereby aligning morphological features along interpretable axes. PCA applied to this domain identifies dominant structural modes—here referred to as morphological base components—which form an orthogonal basis spanning the latent shape space of the class.

Each novel observation is interpreted as a projection into this latent space:

\[
\mathbf{x}_{\text{new}} \approx \mathbf{\bar{x}} + \sum_{k=1}^N c_k \cdot \mathbf{PC}_k
\]

where $\mathbf{\bar{x}}$ is the class mean (in polar coordinates), $\mathbf{PC}_k$ denotes the $k$-th principal component, and $c_k$ are the projection coefficients specific to the observation.

Empirically, reconstructions of unseen test samples confirm that:
\begin{itemize}
    \item The PCA basis functions as a shape prior, constraining reconstructions to plausible morphological configurations.
    \item The learned latent space encodes the class-defining variations with minimal reconstruction loss.
    \item The dimensionality of the latent space is significantly lower than the original image space, yet sufficient to retain structural integrity.
\end{itemize}

This suggests that the latent PCA space serves as a morphological probability manifold, where each observation is resolved as the most probable linear combination of learned modes.

Our approach thus frames class morphology as a constrained probabilistic projection within a linear latent space, induced by PCA over polar-transformed training instances. This representation inherently filters high-frequency sampling artifacts and emphasizes class-consistent structural information.

\begin{figure}[h]
    \centering
    \includegraphics[width=1.0\linewidth]{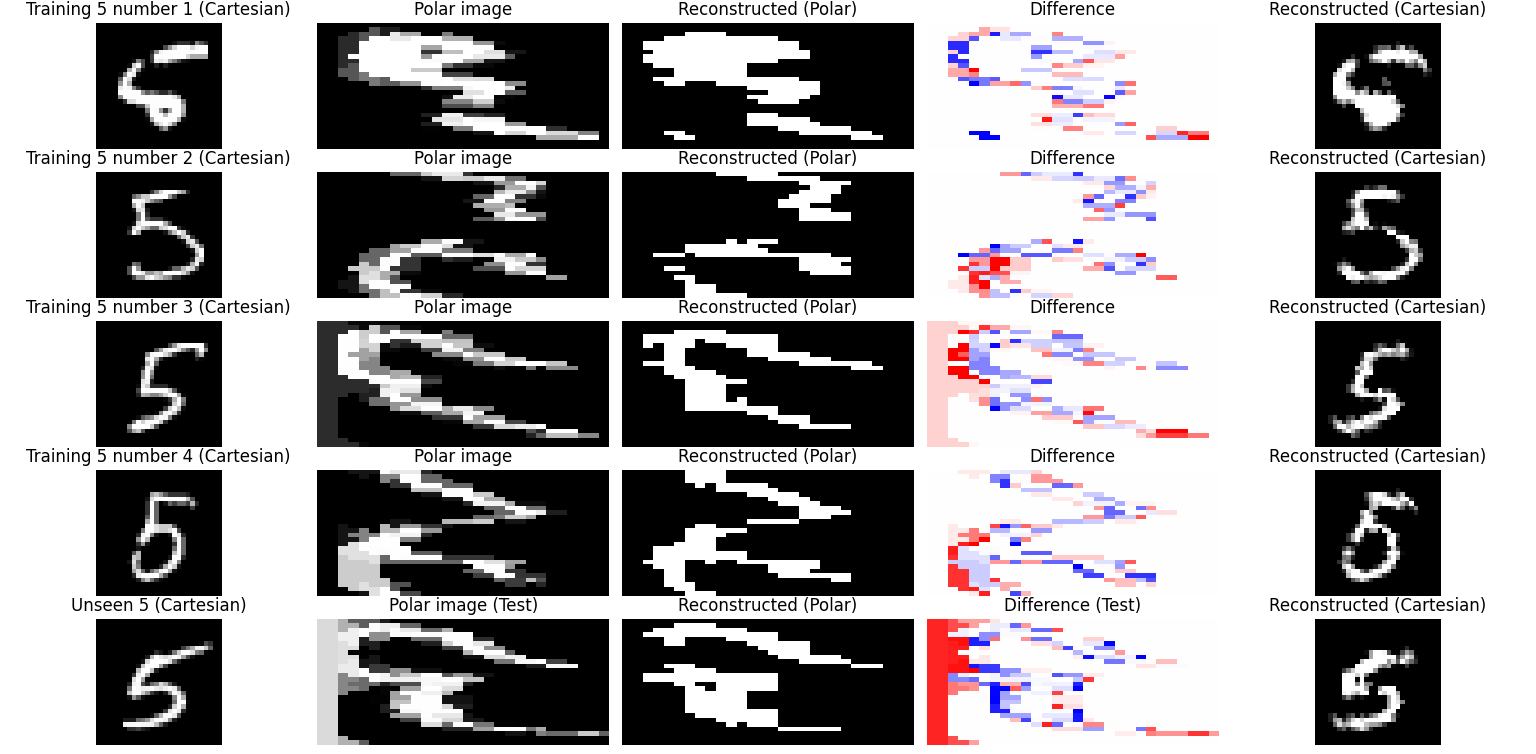}
    \caption{Enter Caption}
    \label{fig:enter-label}
\end{figure}

\section{Numerical Example: Typology Reconstruction via PCA}

We consider a training image \( X_{\text{train}} \in \mathbb{R}^{4 \times 4} \), where each row corresponds to an angular segment, and each column encodes a spatial pattern:

\[
X_{\text{train}} =
\begin{bmatrix}
1 & 1 & 0 & 0 \\
1 & 0 & 0 & 0 \\
1 & 0 & 0 & 0 \\
1 & 0 & 0 & 0 \\
\end{bmatrix}
\]

Applying Principal Component Analysis (PCA) to \( X_{\text{train}} \) yields a component matrix \( W_{\text{train}} \in \mathbb{R}^{4 \times 4} \) and a mean vector \( \mu_{\text{train}} \in \mathbb{R}^{4} \):

\[
W_{\text{train}} =
\begin{bmatrix}
0 & 1 & 0 & 0 \\
0 & 0 & 1 & 0 \\
0 & 0 & 0 & 1 \\
1 & 0 & 0 & 0 \\
\end{bmatrix}, \quad
\mu_{\text{train}} =
\begin{bmatrix}
1.0 & 0.25 & 0.0 & 0.0
\end{bmatrix}
\]

We now introduce a test image \( X_{\text{test}} \) that partially matches the structure of the training set:

\[
X_{\text{test}} =
\begin{bmatrix}
0.3 & 1 & 0 & 0 \\
0.5 & 0 & 0 & 0 \\
0   & 0 & 0 & 0 \\
0   & 0 & 0 & 0 \\
\end{bmatrix}
\]

The test image is projected into the latent PCA space defined by the training components:

\[
Z_{\text{test}} = (X_{\text{test}} - \mu_{\text{train}}) W_{\text{train}}^\top
\]

and is reconstructed via:

\[
\hat{X}_{\text{test}} = Z_{\text{test}} W_{\text{train}} + \mu_{\text{train}}
\]

The resulting reconstruction \( \hat{X}_{\text{test}} \) approximates the original test image in rows where typological features are shared. This simple numerical example demonstrates:

\begin{itemize}
    \item The component matrix \( W_{\text{train}} \) captures a structural typology learned from the training data.
    \item A test instance containing partial patterns from this typology can be reconstructed by projection into the trained PCA basis.
    \item The reconstruction depends on the coherence between \( Z_{\text{test}} \) and the component matrix \( W_{\text{train}} \); both must align structurally.
    \item Complex typologies can be decomposed into simpler sub-typologies encoded in individual PCA components.
\end{itemize}

This analysis supports the use of PCA as a typological representation mechanism, where inference and partial reconstruction are possible when the structural basis is preserved.

This validates the use of PCA as a framework not only for compression or noise filtering, but also for typological inference and reconstruction from partial data. In the context of polar image segmentation, this approach enables the recovery of structured angular features even when partial or noisy.

\begin{figure}[h]
    \centering
    \includegraphics[width=1\linewidth]{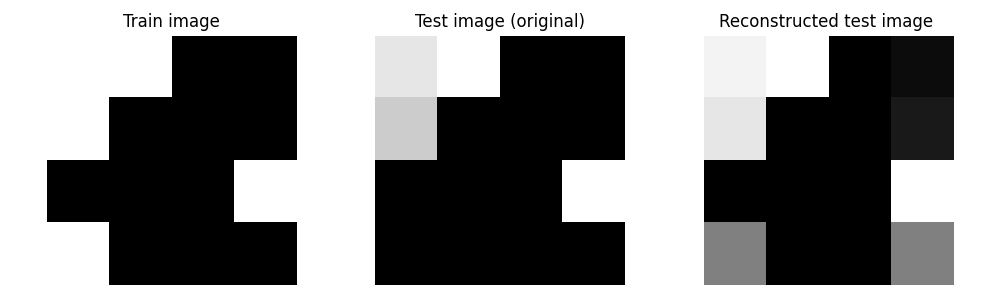}
    \caption{A simple binary block pattern is used as training image (left). The test image (center) contains an additional gray patch in the upper-left corner that deviates from the training set. The reconstructed test image (right) shows that the deviation is retained, indicating that the model captures and preserves novel features during reconstruction.}
    \label{fig:4}
\end{figure}

To evaluate how the reconstruction process handles deviations from the training data, we constructed a minimal example using binary block patterns. A single image was used as the training sample, and a test image was created by introducing a localized variation in the upper-left region. The goal is to observe whether the reconstruction process suppresses or retains such novel features. The result is shown in Figure~\ref{fig:4}.

\section{Experiments and Reflections on Results}

As the theoretical exposition has already illustrated, many of the key mechanisms and intuitions underlying our method are embedded directly into the model design—particularly the use of polar PCA and the decomposition of angular segments relative to a class-mean image. Rather than strictly separating theory from results, we acknowledge that the methodological framework is inherently demonstrative.

In our experiments, we apply this approach by constructing a class-specific polar PCA basis from just two MNIST images per class (i.e., 20 samples in total). Each class-mean polar image is used to define a local component space, which captures dominant angular variations within that digit class. The resulting PCA vectors serve not only as reconstruction tools, but also as manipulable handles for controlled transformations, such as polar rotation or morphological variation.

What is particularly noteworthy is that this method enables us to both generate and manipulate data in a structure-aware manner: the PCA vectors are locally tied to a specific image geometry, yet globally anchored to class-level structure via the average image. In this way, the method inherits the interpretability of eigen-digits while retaining the individual specificity of each instance. To the best of our knowledge, such a combination of local and global PCA structuring over polar-transformed images has not been previously described in the literature.

To evaluate the framework, we trained a lightweight conditional generator to synthesize PCA codes that reconstruct images within the class-specific polar subspace. These reconstructed images are then passed through a frozen teacher model (LeNet-5), which serves as a typological validator. The training process is analogous to a guided search over the PCA space—akin to a game of Battleship—where the generator learns to discover semantically valid configurations through teacher feedback.

Despite the minimalist setup, we were able to synthesize training data that led to surprisingly strong results. Using only two real images per class to define the PCA spaces, and generating all training data synthetically via our framework, we trained a new LeNet-5 model that achieved 69\% accuracy on the real MNIST test set. This demonstrates the potential of low-dimensional, structure-aware generation in data-free learning scenarios.

All relevant code for the PCA generation, polar transformation, and image reconstruction has been made publicly available to encourage replication, critical evaluation, and further development of the method.

\section{Conclusion}

C2G-KD presents a novel generative framework for data-free knowledge distillation, grounded in a topology-first philosophy and enabled by class-specific PCA constraints. By leveraging polar-transformed representations and constructing PCA subspaces from just a few real samples, the method provides a structured latent space where synthetic variation becomes both interpretable and controllable.

Unlike approaches that rely on unconstrained generative modeling, our method incorporates geometric priors derived from class-mean images, enabling the generator to produce data that is topologically plausible by construction. These samples are then validated semantically through a frozen teacher model, ensuring alignment with the typological boundaries defined during training.

The polar PCA strategy—applied segment-wise across angular slices—proved particularly effective, allowing us to reconstruct and manipulate image morphology with minimal representational overhead. By synthesizing and perturbing PCA coefficients rather than full images, we reduce the complexity of the generation task while preserving the structural essence of each class.

In our experiments, this approach enabled the creation of synthetic datasets that successfully trained student models with strong downstream performance—achieving 69\% accuracy on MNIST with only 20 real images used for PCA extraction. These results suggest that substantial gains in efficiency and interpretability can be achieved when model design is informed by the underlying geometry of the data.

We believe this integration of structured generation, semantic feedback, and minimal data dependency represents a promising direction for future research. To support open inquiry and community validation, we have made all core components of the framework publicly available.

\section*{Acknowledgements}
The authors gratefully acknowledge their independence.

\section*{Code Availability}
\url{https://github.com/mnbe1973/C2G-KD}
\bibliographystyle{ieeetr}  % Välj stil, t.ex. IEEE
\bibliography{references}   % Namnet på din .bib-fil (utan filändelse)

\end{document}